\documentclass[letterpaper, 10 pt, journal]{IEEEtran}
\IEEEoverridecommandlockouts                          
\usepackage{graphics} 
\usepackage{epsfig}
\usepackage{times}
\usepackage{amsmath}
\usepackage{amssymb}
\usepackage{threeparttable}
\usepackage{booktabs}
\usepackage{soul}
\usepackage{subfigure}
\usepackage{xcolor}
\usepackage[hidelinks]{hyperref}
\usepackage{algorithm}
\usepackage{comment}
\usepackage{algpseudocode}
\usepackage{float}
\usepackage{xcolor}
\floatstyle{ruled}
\restylefloat{algorithm}
\usepackage{tcolorbox}
\usepackage{multirow}
\usepackage{graphicx}
\usepackage{adjustbox}
\usepackage{censor}
\usepackage{cite}
\usepackage{graphics}

\title{\LARGE \bf
SELF-VLA: A \underline{S}kill \underline{E}nhanced Agentic Vision-\underline{L}anguage-Action \underline{F}ramework for Contact-Rich Disassembly}

\author{Chang Liu$^{1}$, Sibo Tian$^{1}$, Xiao Liang$^{2,*}$, and Minghui Zheng$^{1,*}$
\thanks{This work was supported by the USA National Science Foundation under Grant No. 2422826 and 2527316. Portions of this research were conducted with the advanced computing resources provided by Texas A\&M High Performance Research Computing.}
\thanks{$^{1}$ Chang Liu, Sibo Tian and Minghui Zheng are with the J. Mike Walker '66 Department of Mechanical Engineering, Texas A\&M University, College Station, TX 77843, USA. {\tt\small Emails: {changliu.chris, sibotian, mhzheng}@tamu.edu.}}
\thanks{$^{2}$ Xiao Liang is with the Zachry Department of Civil and Environmental Engineering, Texas A\&M University, College Station, TX 77843, USA. {\tt\small Email: xliang@tamu.edu.}}
\thanks{* Corresponding Authors.}
}

\begin{document}

\maketitle
\begin{abstract}

Disassembly automation has long been pursued to address the growing demand for efficient and proper recovery of valuable components from the end-of-life (EoL) electronic products. Existing approaches have demonstrated promising and regimented performance by decomposing the disassembly process into different subtasks. However, each subtask typically requires extensive data preparation, model training, and system management. Moreover, these approaches are often task- and component-specific, making them poorly suited to handle the variability and uncertainty of EoL products and limiting their generalization capabilities. All these factors restrict the practical deployment of current robotic disassembly systems and leave them highly reliant on human labor. With the recent development of foundation models in robotics, vision-language-action (VLA) models have shown impressive performance on standard robotic manipulation tasks, but their applicability to complex, contact-rich, and long-horizon industrial practices like disassembly, which requires sequential and precise manipulation, remains limited. To address this challenge, we propose SELF-VLA, an agentic VLA framework that integrates explicit disassembly skills. Experimental studies demonstrate that our framework significantly outperforms current state-of-the-art end-to-end VLA models on two contact-rich disassembly tasks. The video illustration can be found \href{https://zh.engr.tamu.edu/wp-content/uploads/sites/310/2026/03/IROS-VLA-Video.mp4}{\textcolor{blue}{here}}.

\end{abstract}

\section{INTRODUCTION}

Electronic waste (e-waste) reached 62 million tonnes globally in 2022, making it one of the fastest-growing waste streams \cite{balde2024global}. However, only 22.3\% of this volume was formally documented as recycled, leaving the majority improperly treated. Besides the environmental issues associated with improper treatment, end-of-life (EoL) electronics contain substantial quantities of rare earth elements (REEs), whose loss further aggravates existing supply chain risks \cite{maani2024disassembly}. Recovering these materials requires disassembly as the first operational step in both recycling and remanufacturing processes. However, collected EoL products (e.g., desktops, phones, and laptops) exhibit significant variability across different brands and considerable uncertainty due to unpredictable usage conditions, requiring complex decision-making and planning during disassembly operations \cite{lee2024review}. Traditionally, manual disassembly has dominated this practice because humans can effectively handle the uncertainty and variability inherent in different EoL products. However, the high labor costs and low efficiency of manual processes cannot meet the increasing demand for EoL desktop disassembly and render it unprofitable \cite{liu2026raise}. Additionally, health concerns during the disassembly operation further motivate the integration of robots to reduce human involvement \cite{andeobu2023informal}. Current robotic-assisted disassembly approaches rely on a regimented sequential pipeline consisting of perception \cite{deng2025learning}, task planning \cite{yu2022disassembly, lee2022task}, motion planning \cite{asif2024robotic}, and manipulation \cite{sliwowski2025reassemble}. While effective in a controlled setting, these task-specific designs rely heavily on explicit modeling and staged engineering, which accumulate errors and struggle under real-time uncertainty.

\begin{figure}[t]
\vspace{0.1 in}
    \begin{center}
        \includegraphics[width=0.45\textwidth]{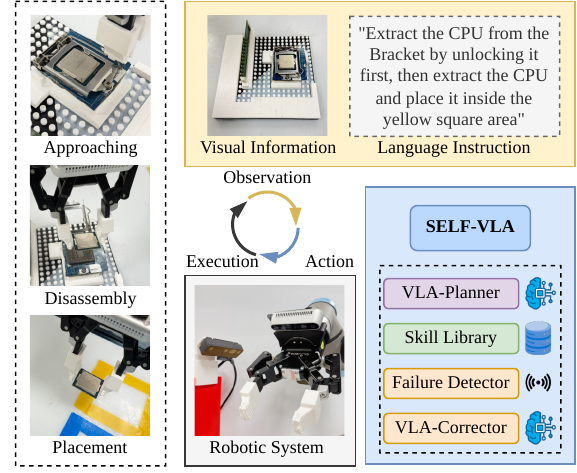}
        \caption{SELF-VLA framework 
        }
    \label{figure_struc} 
    \end{center}
    \vspace{-0.1in}
\end{figure}

Large foundation models have advanced the deployment of artificial intelligence (AI) across different domains \cite{naveed2025comprehensive}. Their pre-training on large-scale multi-modal datasets enables strong generalization. In robotics, vision-language-action (VLA) models, which leverage vision-language foundation models as reasoning backbones to directly generate executable robot actions, have emerged as a promising direction for both industry and academia \cite{kawaharazuka2025vision}. VLA models aim to learn from large-scale human demonstrations, enabling robots to understand perceptual inputs with language instruction and execute the task directly without multiple-stage splits. While current VLA models have shown strong results for robotic tabletop daily tasks \cite{kim2024openvla, kim2025fine, intelligence2025pi_}, their applicability to more complex long-horizon and contact-rich settings remains limited. The training datasets, like LIBERO \cite{liu2023libero} and DROID \cite{khazatsky2024droid}, primarily consist of relatively simple daily life manipulation tasks with loose constraints and lack industry-level operational data. In contrast, industrial environments demand high-precision manipulation in sequential, contact-rich operations. Beyond basic pick-and-place tasks, these scenarios often involve standardized extraction procedures that must adhere to strict operational constraints. Meeting such requirements remains challenging for current VLA models, even after domain-specific fine-tuning \cite{li2025transferring, liu2025vision}. In these end-to-end approaches, the entire task is governed by a single continuous policy without explicitly accounting for distinct stages that impose different execution constraints. As a result, ensuring consistent behavior in long-horizon operations that require strict procedural sequences remains challenging. This raises the question of how to retain the flexibility of end-to-end learning while incorporating stage-specific execution constraints.

Recently, agentic frameworks built upon large language models (LLMs) have attracted increasing attention. These frameworks typically operate through an iterative loop involving task understanding, planning, tool-augmented execution, and feedback-driven evaluation \cite{acharya2025agentic}. Although LLM agents benefit from retrieval-augmented generation (RAG) and external tool integration, their reasoning remains largely prompt-driven and lacks structured procedural control, limiting their effectiveness in long-horizon tasks with strict constraints. The recent emergence of LLM agent skills addresses this limitation by formalizing procedural knowledge into structured and reusable workflows that explicitly guide both reasoning and execution \cite{xu2026agent}. This structured agentic paradigm motivates extending the applicability of VLA models in robotics. Similar to LLM agents, VLA systems operate over multi-modal inputs and long-horizon objectives under operational constraints. Introducing skills provides a structured mechanism for organizing task execution, particularly in stages that require precise and sequential operations.

Motivated by this agentic perspective, we propose SELF-VLA, a VLA framework with explicit skill integration for robotic disassembly tasks, as shown in Figure~\ref{figure_struc}. The framework consists of three coordinated components: a VLA-planner, a skill library, and a VLA-corrector. The VLA-planner interprets language instructions and visual observations to understand the task and generate continuous control actions toward the target component. Once the planner outputs a stop token indicating that the robot has reached a suitable state for contact-rich manipulation, the framework invokes the corresponding skill from the skill library. The skill executes the manipulation along a predefined waypoint sequence, with integrated grasp and drop detection. If a failure is detected, the VLA-corrector is activated to regrasp the component and reinvoke the skill. The main contributions of this work can be summarized as follows:
\begin{itemize}
    \item We propose SELF-VLA, a skill enhanced agentic VLA framework that integrates structured disassembly skills with failure recovery for long-horizon contact-rich manipulation.
    \item We evaluate the proposed framework on two robotic disassembly tasks and compare it with end-to-end VLA baselines. Experimental results demonstrate that SELF-VLA significantly improves task success rates.
\end{itemize}

The remainder of the paper is organized as follows. Section 2 reviews related work on robotic disassembly and VLA models. Section 3 presents the SELF-VLA framework, including the framework design, dataset construction, and fine-tuning procedures. Section 4 describes the experimental setup, evaluations, and detailed result analysis. Section 5 concludes the paper.

\begin{figure*}[htbp]
    \begin{center}
        \includegraphics[width=0.95\textwidth]{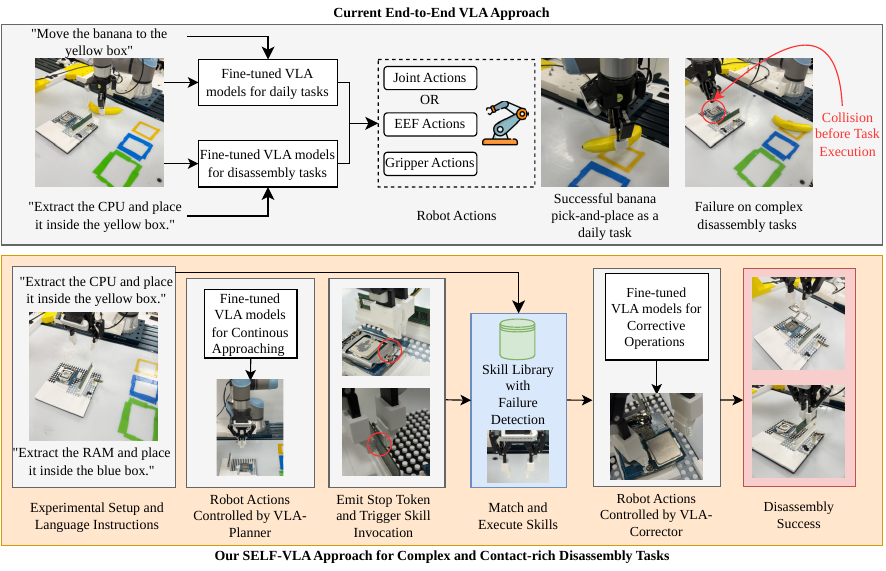}
        \caption{Comparison between current VLA approaches and our SELF-VLA framework.}
    \label{figure_framworkcom} 
    \end{center}
    \vspace{-0.1in}
\end{figure*}

\section{RELATED WORK}

\subsection{Robotic Disassembly}

Current robotic-assisted disassembly approaches rely on a sequential pipeline, which requires the robot to perceive and process multi-modal information (e.g., visual information, CAD data, product information, and environmental information) to generate a feasible plan for robot actions at different stages \cite{lee2024review}. Prioli et al. \cite{prioli2022disassembly} extract interference matrices from CAD files through collision tests to automatically build a precedence matrix, providing collision-free disassembly sequences for robotic task planning. Zhang et al. \cite{zhang2023automatic} propose a YOLOv4-based framework to detect and localize screws from visual inputs, recommending matched tools for robotic unscrewing tasks.
However, such information is usually incomplete or unavailable in real-world scenarios, limiting the practical feasibility of these algorithms \cite{foo2022challenges}. Liu et al. \cite{liu2026raise} proposed an automated selective disassembly system for EoL phones, achieving high-throughput under structured operating conditions. However, it relies on the specific phone structures with limited generalization capability.

Human-robot collaborative disassembly (HRCD) has been proposed as an intermediate solution that integrates humans' dexterous manipulation and flexible decision-making with robots' reliability, repeatability, and precision \cite{hjorth2022human, lee2022task}. Lee et al. \cite{lee2022task} proposed a resource-constrained optimization algorithm to perform task planning for HRCD, reducing the operation time while ensuring safety. However, the involvement of humans brings additional uncertainty for robots to do motion planning, particularly when predicting human actions to guarantee safety. Tian et al. \cite{tian2025real} distill a diffusion-based motion predictor into a lightweight one-step MLP model for real-time stochastic 3D human motion prediction in HRCD. While recent research has demonstrated impressive results, HRCD in a narrow workspace for EoL electronic disassembly tasks can bring hesitation and trust issues between humans and robots, which may lower the operational efficiency and limit throughput \cite{hopko2023physiological}. 
Additionally, HRCD still largely preserves the original sequential workflow, relying on human involvement to compensate for the limited real-time decision-making and dexterous manipulation capabilities, while leaving the structured task-dependent stages and accumulated errors unaddressed. An automated and data-driven disassembly system still remains as a long-term objective \cite{foo2022challenges}.

\subsection{Vision-Language-Action Models}
Vision-Language-Action (VLA) models directly map visual observations and language instructions to robot actions. These models differ in how robot actions are represented and generated, and existing approaches adopt distinct strategies for learning continuous control signals from multi-modal inputs. OpenVLA \cite{kim2024openvla} discretizes robot actions into tokens and predicts single-step actions autoregressively using a large language model backbone. OpenVLA-OFT \cite{kim2025fine} replaces tokenized actions with a continuous regression head and introduces action chunking to predict short-term future sequences. The predicted action chunk is executed sequentially before the next inference step, resulting in a higher control frequency with smooth robot motion. $\pi_{0.5}$ \cite{intelligence2025pi_} differs by employing a flow matching action head that generates action chunks through iterative denoising, enabling expressive multi-modal action distributions. Octo \cite{team2024octo} adopts a conditional diffusion policy head to model continuous action chunks through multi-step denoising, allowing it to capture action distributions while maintaining precise control.

While VLA models integrate multi-modal perception and robot action in an end-to-end manner, their applicability in real-world applications remains challenging \cite{din2025vision}. Liu et al. \cite{liu2025vision} conduct a case study evaluating different VLA models in contact-rich disassembly tasks. Even after fine-tuning, the models achieved near zero success, in contrast to their performance on everyday manipulation tasks commonly represented in VLA training datasets. The failures were mainly attributed to out-of-distribution (OOD) inputs and unintended collisions occurring after contact. Lu et al. \cite{lu2025vla} propose VLA-RL, an online reinforcement learning framework to improve autoregressive OpenVLA policy beyond offline demonstrations training, thereby mitigating distributional shift and improving robustness to out-of-distribution manipulation scenarios. Yang et al. \cite{yang2025agentic} propose a standardized action procedure (SAP) to control robots in an agentic way, including a reasoning LLM model to plan the tasks, a VLA model as executor, and a vision-language model (VLM) for verifier to form a closed loop for long-horizon manipulation. Zhao et al. \cite{zhao2025vla} propose an agentic extension for the OpenVLA framework, integrating task planning, web search, object grounding, and memory-based retrieval to improve the concept-level generalization in robotic manipulation. While they substantially improve the performance on the LIBERO benchmark by introducing the agentic framework, it does not explicitly address precision-critical manipulation constraints. Pang et al. \cite{pang2026sci} introduce a Sci-VLA inference plugin that inserts LLM-generated trajectories between atomic tasks, using semantic retrieval to estimate target start poses and restore task-to-task coherence during execution. Shi et al. \cite{shi2025memoryvla} propose a MemoryVLA framework, integrating a working memory and a perceptual-cognitive memory bank to VLA models to capture and save context for long-horizon manipulation tasks. 

\section{METHODOLOGIES}

\subsection{SELF-VLA}
In our proposed agentic framework, the VLA operates as an agent that perceives the environment, generates low-level robot action commands, and determines when to invoke the skill library to perform precise, contact-rich operations that exceed its learned manipulation capabilities. The comparison between the SELF-VLA framework and the current end-to-end approach is presented in Figure~\ref{figure_framworkcom}. The framework consists of three components: a VLA-planner, a skill library, and a VLA-corrector. The VLA-planner interprets the language instruction $l$ and visual observations $o_t$ to generate continuous action commands $a_t$ that navigate the robot toward the target component, and outputs a stop token $g_t$ to initiate skill execution. The skill library then executes the complete contact-rich disassembly sequence, including component extraction and placement, with failure detection after the pick-up action. If a failure is detected, the VLA-corrector intervenes to recover the dropped component and re-invokes the skill to complete the remaining task. The implementation of the SELF-VLA framework is illustrated in Algorithm~\ref{alg:self_vla}.

The VLA-planner $\pi_\theta$ outputs 6-DoF end-effector or joint actions along with a gripper action at each time step. Since the VLA architecture produces actions in a fixed output dimension, introducing a discrete stop signal would require redesigning the output head and retraining from scratch. Instead, we encode the stop token within the gripper action dimension by assigning a value $g_{\text{stop}}$, and $g_{\text{stop}}=255$ in our case, which exceeds the physical range of the gripper and does not appear during normal operation. During deployment, skill execution is triggered once the planner outputs this value $g_t = g_{\text{stop}}$, enabling the planner to autonomously decide when to transition from directly controlling the robot to invoking the skill library for the subsequent manipulation.

Invoked by the stop token generated by the VLA-planner, each skill $k$ in the library is defined as a sequence of waypoints $\tau$ generated by $\mathcal{G}_k$ with two coordinate types: relative movement $\tau_{rel}$ from the current tool center point (TCP) pose for the extraction phase, and absolute positions $\tau_{abs}$ in the robot base frame for the placement phase. The relative waypoints allow the skill to adapt to varying start poses determined by the VLA-planner, while the absolute waypoints are used for fixed goal positions, such as the different component placement locations. The path blending radius is applied at each waypoint to generate smooth trajectories, with a smaller radius at the initial waypoints for precise positioning. The failure detection is performed after the pick-up action by comparing the gripper width difference between control commands and observations. If the difference falls below a predefined threshold, indicating that the gripper failed to secure the component, the VLA-corrector is triggered. The CPU extraction skill and the RAM removal skill consist of 23 and 8 waypoints, respectively. 

\begin{algorithm}[t]
\small
\caption{SELF-VLA}
\label{alg:self_vla}
    \begin{algorithmic}[1]
        \Require Robot state $s_t$, observation $o_t$, instruction $l$
        \Require Planner $\pi_\theta$, Corrector $\pi_\phi$, skill generator $\mathcal{G}_k$, skill controller $\kappa_k$, stop token $g_{\text{stop}}$
        \Require Transition $\mathcal{F}$, observation function $h$, failure detector $\textsc{Fail}(\cdot)$
        \State $\mathbf{e}\gets f_l(l)$,\quad $k\gets\arg\max_k\psi_k(\mathbf{e})$ \Comment{\textcolor{gray}{instruction encoding \& skill selection}}
        \State mode $\gets$ planner \Comment{\textcolor{gray}{control mode $\in\{$planner, skill, corrector$\}$}}
        \While{task not completed}
            \State $\mathbf{z}_t\gets f_o(o_t)$ \Comment{\textcolor{gray}{visual observation encoder}}
            \State $a_t=[u_t,g_t]\gets
                \begin{cases}
                \pi_\theta(\mathbf{z}_t,\mathbf{e}) & \text{if planner}\\
                \kappa_k(s_t,\tau) & \text{if skill}\\
                \pi_\phi(\mathbf{z}_t,\mathbf{e}) & \text{if corrector}
                \end{cases}$ 
            \State $s_{t+1}\gets\mathcal{F}(s_t,a_t)$,\; $o_{t+1}\gets h(s_{t+1})$ \Comment{\textcolor{gray}{robot execution and observation update}}
            \If{$g_t=g_{\text{stop}}$}
                \State $\tau\gets\mathcal{G}_k(s_{t+1})$, mode $\gets$ skill \Comment{\textcolor{gray}{invoke skill}}
            \EndIf
            \If{$\textsc{Fail}(o_{t+1},a_t)$}
                \State mode $\gets$ corrector \Comment{\textcolor{gray}{failure detector}}
            \EndIf
            \If{$g_t=g_{\text{stop}}$ \textbf{and} mode$=$corrector}
                \State $\tau\gets\mathcal{G}_k^{rem}(s_{t+1})$, mode $\gets$ skill \Comment{\textcolor{gray}{resume}}
            \EndIf
            \State $s_t\gets s_{t+1}$,\; $o_t\gets o_{t+1}$ \Comment{\textcolor{gray}{advance system state}}
        \EndWhile
    \end{algorithmic}
    \end{algorithm}

The VLA-corrector $\pi_\phi$ is autonomously activated by the SELF-VLA under two conditions: when the failure detection identifies an unsuccessful grasp, or when the continuous drop detection identifies a component loss during the subsequent skill execution after an initially successful grasp, which monitors the gripper width at 50 Hz during the placement phase. Once triggered, the VLA-corrector receives the same input modalities as the VLA-planner to generate action commands and control the robot toward the component and pick it up again. It then emits the same stop token to re-invoke the skill, which resumes from the absolute waypoints $\tau_{abs}$ to complete the placement operation, as the extraction phase has already been completed.

\begin{figure*}[htbp]
    % \vspace{-0.1in}
    \begin{center}
        \includegraphics[width=0.95\textwidth]{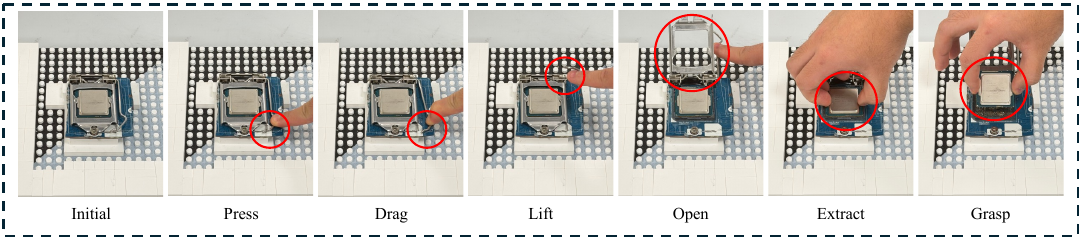}
        \caption{The complexity of the CPU extraction task in EoL desktop disassembly.}
    \label{figure_cpu} 
    \end{center}
    \vspace{-0.1in}
\end{figure*}

\subsection{Dataset Construction}
To train the agentic framework and the end-to-end VLA baselines, we construct a real-world disassembly dataset for high-value component extraction from EoL desktops in two stages. In the first stage, the skill library is constructed by teleoperating the robot through the contact-rich manipulation operations of the disassembly sequence and recording the waypoints for each skill. In the second stage, we collect full disassembly demonstrations in which the operator remotely controls the robot to approach the target component, trigger the skill, and provide corrections when the grasp fails.  

To introduce more variance to the dataset, the CPU socket and RAM slot are placed at 8 different locations and orientations on the fixture board. For each configuration, the robot starts from a fixed home position, and the operator controls it to approach the target component. Once the end-effector reaches the approximate start position, the operator triggers the skill. If no failure is detected, the robot completes the remaining skills to complete the placement. If failure is detected, the operator provides corrective input through teleoperation to pick up the component before the skill resumes placement. The idle time between the robot's pause from failure detection and the operator's correction input is excluded from the recording. 

To enable separate training for VLA-planner, VLA-corrector, and the end-to-end baseline, each demonstration episode is labeled into four phases: approaching, skill execution, correction, and skill resumption. 
The approaching phase spans from the initial home position to the point where the skill is triggered. The skill execution phase covers the full skill manipulation operation, from trigger to placement completion or failure detection. The correction phase captures the operator's recovery input after a detected failure, followed by the skill resumption phase, which completes the remaining placement. During the data processing, a stop action is designed as three consecutive frames with zero delta movement and a gripper value of 255, signaling the transition from VLA control to skill invocation. These repeated frames are appended to the end of each approaching and correction trajectory to prepare the training data for VLA-planner and VLA-corrector. The data in the approaching phase is used to train the VLA-planner, the correction phase data, supplemented with the pick-and-lift actions from the skill execution phase, trains the VLA-corrector, and the full episodes without stop actions are used to train the end-to-end baseline VLA models.

\subsection{Fine-tuning Process}
We utilize LoRA to fine-tune three well-known base VLA models: OpenVLA \cite{kim2024openvla}, OpenVLA-OFT \cite{kim2025fine}, and $\pi_{0.5}$ \cite{intelligence2025pi_}. For $\pi_{0.5}$ models, we fine-tune both the base $\pi_{0.5}$ model and its expanded version $\pi_{0.5}$-Droid, which is fine-tuned on the DROID dataset \cite{khazatsky2024droid} with massive real-world manipulation demonstrations. 
For each of the four models, we train a VLA-planner, a VLA-corrector, and an end-to-end baseline, each on both the original 30 Hz data and a downsampled 10 Hz variant to evaluate their performances under different data sampling rates. Data from different disassembly tasks are combined for training, with each task specified by its corresponding language instruction.

\section{EXPERIMENTS AND TEST RESULTS}

\begin{figure*}[htbp]
    \vspace{-0.1in}
    \begin{center}
        \includegraphics[width=0.98\textwidth]{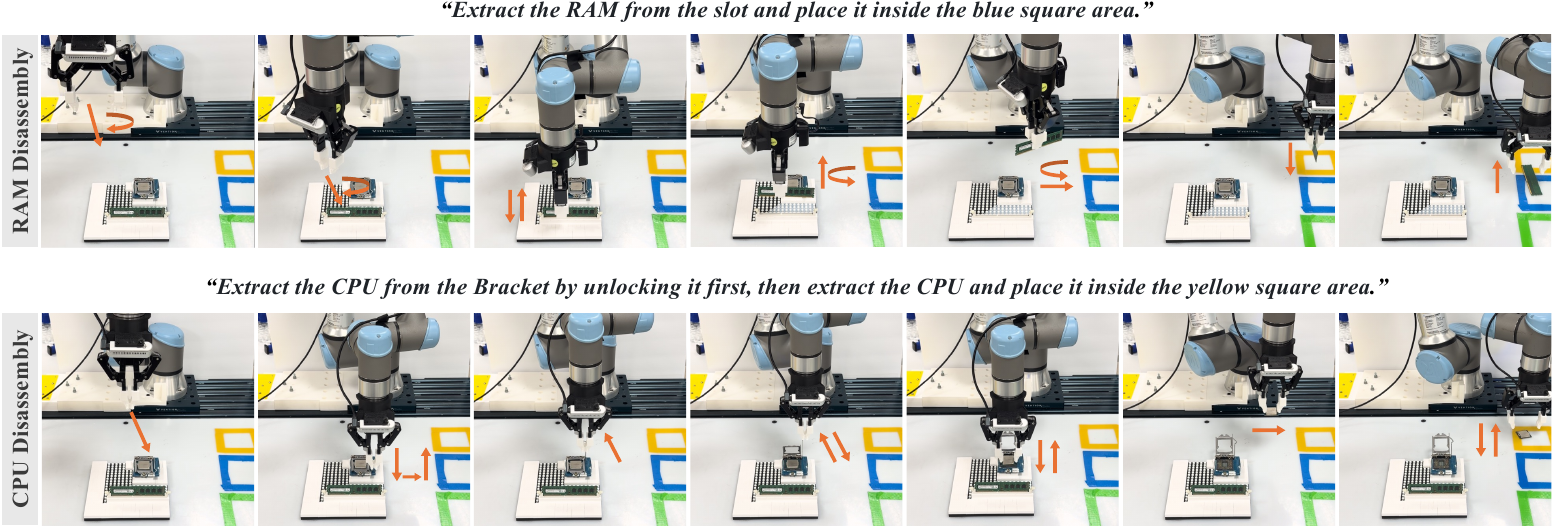}
        \caption{Experimental Studies. Top figures show the results for RAM disassembly, and bottom figures demonstrate CPU disassembly case.}
    \label{figure_demo} 
    \end{center}
    \vspace{-0.1in}
\end{figure*}

\subsection{Environmental Setup}

We select two disassembly tasks from EoL desktops: CPU extraction and RAM module removal. CPU extraction is a multi-step operation that requires unlocking the retention lever, lifting the bracket, and picking up the CPU from the socket, as shown in Figure~\ref{figure_cpu}. RAM module removal requires extracting the module vertically from the tightly constrained slot, where its thin profile and limited clearance demand accurate alignment to prevent damage. These two components also represent two of the most valuable parts in a typical EoL desktop.

The experimental platform consists of a UR5e robot arm from Universal Robots equipped with a Robotiq 2F-85 gripper. We design a customized gripper extension to enable robust disassembly across diverse tasks, particularly for CPU bracket unlocking and CPU extraction steps that require contact-rich operations at specific angles. To reduce environmental complexity, we design a fixture board that securely mounts the CPU socket and the RAM slot. This setup transforms the disassembly task from a compact internal process into a tabletop operation, better aligning it with the task distribution of existing pre-training datasets for VLA models. 

\begin{table*}[t]
    \caption{Comparison of success rates at different stages between our SELF-VLA framework and the end‑to‑end baselines.}
    \centering
    \renewcommand{\arraystretch}{1.1}
    \normalsize
    \resizebox{0.99\textwidth}{!}{
    \begin{threeparttable}
    \begin{tabular}{cccccc ccc ccc ccc}
        \toprule
        \multirow{2}{*}{Task} & \multirow{2}{*}{Method} & \multirow{2}{*}{Stage}
        & \multicolumn{3}{c}{OpenVLA}
        & \multicolumn{3}{c}{OpenVLA-OFT}
        & \multicolumn{3}{c}{$\pi_{0.5}$}
        & \multicolumn{3}{c}{$\pi_{0.5}$-Droid} \\
        \cmidrule(lr){4-6}
        \cmidrule(lr){7-9}
        \cmidrule(lr){10-12}
        \cmidrule(lr){13-15}
        & & & PT & FT-10Hz & FT-30Hz
        & PT & FT-10Hz & FT-30Hz
        & PT & FT-10Hz & FT-30Hz
        & PT & FT-10Hz & FT-30Hz \\
        \midrule
        \multirow{6}{*}{RAM Removal}
        & \multirow{3}{*}{End-to-End} & Approaching 
            & 0/20 & 0/20 & 0/20  & 0/20 & 3/20 & 2/20  & 0/20 & 15/20 & 14/20  & 0/20 & 17/20 & 15/20 \\[-0.3ex]
        &  & Disassembly 
            & 0/20  & 0/20  & 0/20   & 0/20  & 0/20  & 0/20   & 0/20  & 6/20 & 4/20 & 0/20  & 9/20 & 7/20 \\[-0.3ex]
        &  & Final Success 
            & 0/20  & 0/20  & 0/20   & 0/20  & 0/20  & 0/20   & 0/20  & 4/20 & 3/20  & 0/20  & 7/20 & 5/20 \\[-0.3ex]
        \cmidrule(lr){2-15}
        & \multirow{3}{*}{SELF‑VLA} & Approaching 
            & -   & 0/20 & 0/20  & -   & 7/20 & 4/20  & -   & 11/20 & 8/20 & -   & 12/20 & 11/20 \\[-0.3ex]
        &  & Disassembly 
            & -   & 0/20  & 0/20   & -   & 4/20  & 2/20   & -   & 9/20 & 8/20  & -   & 12/20 & 11/20 \\[-0.3ex]
        &  & Final Success  
            & -   & 0/20  & 0/20   & -   & 4/20  & 2/20   & -   & 9/20 & 8/20  & -   & \textbf{12/20} & 11/20 \\[-0.3ex]
        \midrule
        \multirow{6}{*}{CPU Extraction}
        & \multirow{3}{*}{End-to-End} & Approaching 
            & 0/20 & 0/20 & 0/20  & 0/20 & 17/20  & 18/20   & 0/20 & 18/20 & 15/20  & 0/20 & 18/20 & 17/20 \\[-0.3ex]
        &  & Disassembly
            & 0/20  & 0/20  & 0/20   & 0/20  & 0/20 & 1/20  & 0/20  & 2/20& 0/20 & 0/20  &  1/20& 2/20 \\[-0.3ex]
        &  & Final Success 
            & 0/20  & 0/20  & 0/20   & 0/20  & 0/20  & 0/20   & 0/20  & 0/20 & 0/20  & 0/20  & 1/20 & 2/20 \\[-0.3ex]
        \cmidrule(lr){2-15}
        & \multirow{3}{*}{SELF‑VLA} & Approaching
            & -   & 0/20 & 0/20  & -   & 14/20  & 4/20  & -   & 13/20 & 10/20 & -   & 19/20 & 11/20 \\[-0.3ex]
        &  & Disassembly
            & -   & 0/20  & 0/20   & -   & 10/20  & 1/20   & -   & 11/20 & 6/20  & -   & 17/20 & 8/20 \\[-0.3ex]
        &  & Final Success 
            & -   & 0/20  & 0/20   & -   & 10/20  & 1/20   & -   & 11/20 & 6/20  & -   & \textbf{17/20} & 7/20 \\[-0.3ex]
        \bottomrule
    \end{tabular}
    \begin{tablenotes}
    \normalsize
    \item[*] PT denotes pre-trained models without task-specific fine-tuning.
        FT-10Hz and FT-30Hz denote models fine-tuned with a downsampling rate of 10Hz and 30Hz, respectively. Each reported success rate is computed over 20 independent trials. Approaching means the policy can guide the robot to reach a valid pre-grasp configuration and initiate the contact-rich process. For SELF-VLA, an approaching success corresponds to the planner reaching the target configuration and emitting the stop token to invoke the skill library. Disassembly indicates that the robot can unlock, extract, and grasp the component. Final success reports the number of trials that complete placement, including cases that initially fail but are subsequently recovered by the models (e.g., VLA-corrector, end-to-end models). 
    \end{tablenotes}
    \end{threeparttable}
    }
    \label{tab:overall_success}
\end{table*}

The setup uses an Intel RealSense D435i as a wrist camera that is mounted on the end-effector with a $30^\circ$ incline, providing a clear view of the manipulation area. It also uses an OAK-D Pro as a fixed side camera on a designed structure that is attached to the robot base, preventing potential position drift between the side camera and the robot base that may introduce additional variance across the collected dataset. Both cameras capture RGB images at 1280$\times$720 resolution, which are subsequently resized to 256$\times$256 for storage. Capturing at high resolution prior to downsampling preserves fine visual details while maintaining compatibility with the VLA model input resolution requirements used in this study. All camera parameters are fixed throughout the collection process to ensure image consistency across episodes. After evaluating various teleoperation methods, we use a dual-joystick configuration to provide precise, smooth motion commands to the robot. The image streams from both cameras and the robot state data are recorded synchronously at 30 Hz for alignment purposes. The dataset contains 264 episodes of CPU extraction and 264 episodes of RAM removal, totaling 528 demonstrations.

\subsection{Evaluation}

\begin{figure*}[htbp]
    \begin{center}
        \includegraphics[width=0.98\textwidth]{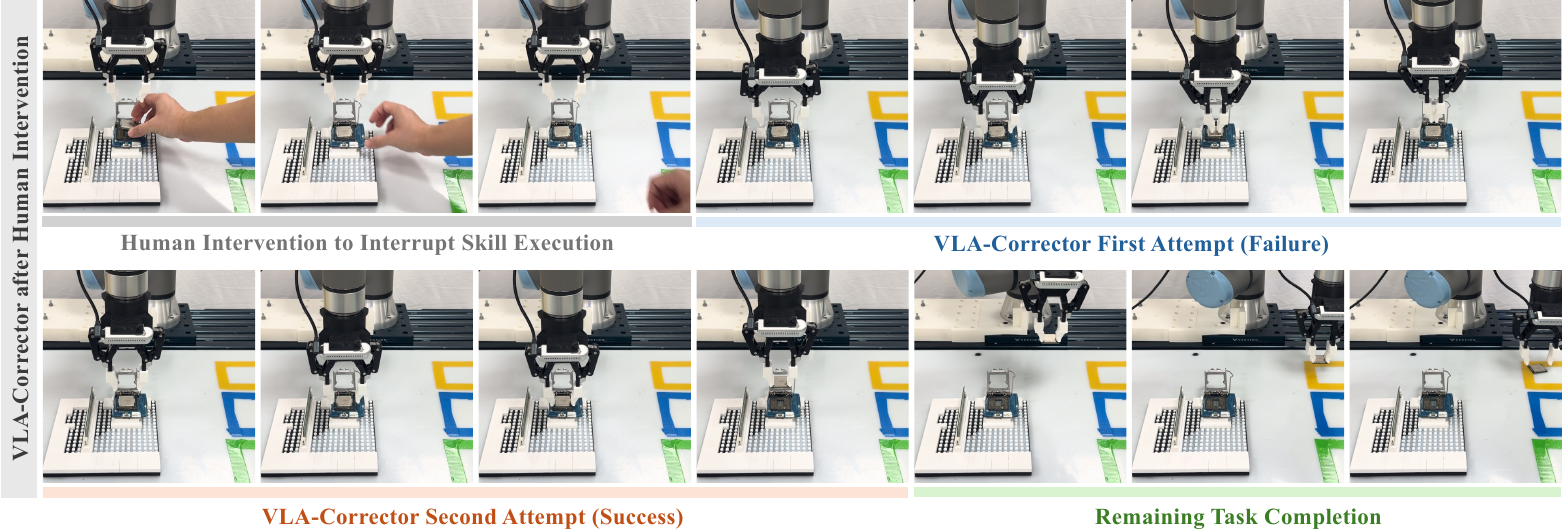}
        \caption{Experimental Study of the VLA-corrector: A human interrupts the skill execution, after which the VLA-corrector attempts to pick the CPU up again and complete the remaining task.}
    \label{figure_corrector} 
    \end{center}
    \vspace{-0.1in}
\end{figure*}

We evaluate our method and all the baselines on both RAM removal and CPU extraction tasks over 20 independent trials. 
For each task, five different component configurations are tested, including positions and orientations on the fixture board that differ from those in the original collected data, with the same robot initial pose. A trial is considered successful only if the robot completes the whole disassembly process and places the extracted component at the desired location. If a failure occurs or the robot moves fully away from the components, the results are recorded, and the test is reset before the next attempt. We compare two deployment strategies. In the end-to-end setting, a single VLA policy controls the whole task sequence. In contrast, the proposed agentic framework is evaluated in a complete configuration, comprising the VLA-planner, skill library, and VLA-corrector. All module interactions, including skill invocation, failure detection, and recovery, are handled automatically without manual intervention. For both strategies, we evaluate four base VLA models, including OpenVLA, OpenVLA-OFT, $\pi_{0.5}$, and $\pi_{0.5}$-Droid, under pre-trained and fine-tuned settings. Additionally, to assess the effect of data downsampling on task performance, we compare results for models that are fine-tuned on data downsampled to 10 Hz and 30 Hz. Pretrained models are evaluated only in the end-to-end setting, as they cannot generate the stop signal required for skill invocation. The evaluation results, including the overall success rates on two selected disassembly tasks, are shown in Table~\ref{tab:overall_success}.

\subsection{Results}
End-to-end baselines achieve no success in the pre-trained setting on either task. Fine-tuning improves performance for some backbones, yet overall success remains limited. In contrast, the SELF-VLA framework consistently increases task completion rates. On average, it improves success by 17\% on RAM module removal and 31\% on CPU extraction. However, the magnitude of improvement depends heavily on the backbone models. OpenVLA does not complete either task under any configuration, including the SELF-VLA setting. Although it maintains appropriate gripper states, it consistently stops several centimeters away from the target components, preventing successful contact. OpenVLA-OFT shows a strong improvement on CPU extraction under the SELF-VLA framework but achieves only marginal success on RAM removal, where positioning errors and collisions frequently occur. Within the $\pi_{0.5}$ family, the fine-tuned models demonstrate substantial gains under the SELF-VLA setting, with an average improvement of 37\%. The $\pi_{0.5}$-Droid variant achieves the highest overall performance, showing an 80\% increase in CPU extraction compared with its end-to-end baseline. The execution sequence of the SELF-VLA framework on the RAM and CPU disassembly tasks is shown in Figure~\ref{figure_demo}. The recovery capability of the VLA-corrector is further demonstrated in Figure~\ref{figure_corrector}.

\begin{figure}[htbp]
    \vspace{-0.1in}
    \begin{center}
        \includegraphics[width=0.4\textwidth]{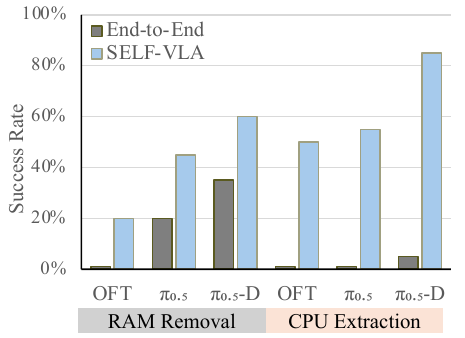}
        \caption{Comparison of final task success rates at FT-10 Hz for three different VLA models OpenVLA-OFT, $\pi_{0.5}$, and $\pi_{0.5}$-Droid under the end-to-end approach and the proposed SELF-VLA framework on selected disassembly tasks. }
    \label{fig:SuccessRate} 
    \end{center}
    \vspace{-0.1in}
\end{figure}

Regarding data sampling rate, models fine-tuned on 10 Hz data often outperform those trained at 30 Hz. Most successful configurations are observed under the 10 Hz setting. Under matched evaluation configurations, SELF-VLA performs better when fine-tuned with 10 Hz data than with 30 Hz data. In several cases, the 30 Hz model fails while the 10 Hz model completes the task, suggesting that 30 Hz data may provide weaker stepwise supervision, as consecutive frames exhibit smaller state changes during training. The failure analysis also reveals a phase-dependent effect. Under end-to-end deployment, the 30 Hz model fails more frequently during the approaching and extraction stages, where precise spatial alignment is required. In contrast, during the unlocking operation, which involves fine contact adjustments, the 30 Hz model shows slightly more stable behavior than its 10 Hz counterpart. This suggests that training on higher-frequency data may benefit fine-grained manipulation, whereas it is less effective for broader motion stages that require spatial alignment. Although this factor influences performance, the structural difference between end-to-end and SELF-VLA deployment contributes the most to the overall improvement. We compare the overall task success rates of models fine-tuned at 10 Hz under the end-to-end baseline and the proposed SELF-VLA framework, as shown in Figure~\ref{fig:SuccessRate}. The OpenVLA model is excluded from the figure since it shows no successful executions under either architecture.

The performance gap between the baseline and the SELF-VLA framework is most obvious on CPU extraction, which requires strictly sequential operations. In the end-to-end setting, the backbones may complete the initial unlocking step but often fail in subsequent stages, such as opening the bracket or grasping the CPU. The SELF-VLA framework changes this pattern by executing these sequential stages through skills, leading to a higher rate of full task completion. The RAM removal task shows a different trend. Since it involves a shorter sequence with tighter geometric tolerance, the main difference lies in when the system initiates the removal process. The robot driven by the end-to-end policy frequently attempts to grasp either too late or with slight misalignment. However, the SELF-VLA framework can emit a stop token to provide a clearer signal to initiate the removal process. With only the extraction point difference, the improvement on RAM removal is therefore smaller and more localized than on CPU extraction. 

We further compare execution time over successful trials. For CPU extraction, SELF-VLA completes the task in 63 seconds on average, compared to 136 seconds for the end-to-end baseline. For RAM removal, the average times are 45 and 50 seconds, respectively. The difference is more evident in the CPU task, which involves multiple ordered stages. By enforcing structured execution, the proposed framework improves operational efficiency rather than merely shortening trajectories. 

Although the environment is simplified and controlled in this setup, OpenVLA and OpenVLA-OFT still fail to complete the tasks under end-to-end deployment. The difficulty, therefore, cannot be attributed solely to environmental complexity. Instead, it reflects limitations in the models' ability to generate accurate actions for precise manipulation. When the policy fails to bring the robot into a sufficiently accurate pre-contact pose, the SELF-VLA framework has little room to improve the remaining stages of the task. This pattern is evident across different backbone models. OpenVLA rarely reaches such a state with its single-camera input, and therefore shows no gain under the SELF-VLA setting. In contrast, the $\pi_{0.5}$ models produce more accurate actions toward the target, enabling skills to be executed from the desired pose and improving the completion of the entire disassembly operation. Among them, $\pi_{0.5}$-Droid performs best overall, likely benefiting from its large-scale real-world manipulation pretraining, which provides stronger prior experience with contact-rich settings. During the evaluation, we also observe out-of-distribution (OOD) failures. When the component orientation differs from that in the collected dataset, the policy often fails to reach a correct pre-contact configuration. These observations indicate that while the genetic framework improves stability within familiar configurations, it does not fully address generalization to unseen spatial variations.

\section{CONCLUSIONS}

Conventional end-to-end VLA models struggle to reliably complete dexterous, contact-rich, high-precision manipulation tasks. In this paper, we present SELF-VLA, a skill enhanced agentic VLA framework for robotic disassembly. By establishing a structured skill library, the robot is endowed with reusable capabilities for complex object manipulation. Combined with a VLA-planner and a VLA-corrector, the proposed agent system can automatically switch between specialized sub-modules for task understanding and execution. We evaluate the proposed agentic setting using four widely adopted VLA models, and experimental results demonstrate that the SELF-VLA framework significantly improves task success rates in robotic disassembly with the help of an established manipulation skill library.

\bibliographystyle{IEEEtran}
\bibliography{ref}{}

\end{document}